\documentclass[a4paper, 10pt, conference]{ieeeconf}      
\usepackage{FG2017}
\usepackage{graphicx}
\usepackage{amsmath,amssymb} 
\usepackage{color}
\usepackage{subfigure}
\usepackage{comment}
\usepackage{authblk}
\usepackage{hyperref}
\usepackage{microtype}
\usepackage{caption}

\FGfinalcopy 

\IEEEoverridecommandlockouts                              

\title{\LARGE \bf
Recurrent Human Pose Estimation
}

\author{Vasileios Belagiannis and Andrew Zisserman\\
Visual Geometry Group\\Department of Engineering Science\\
University of Oxford, UK\\
{\normalsize \textit{vb,az@robots.ox.ac.uk}}}

\begin{document}

\ifFGfinal
\thispagestyle{empty}
\pagestyle{empty}
\else
\author{Anonymous FG 2017 submission\\-- DO NOT DISTRIBUTE --\\}
\pagestyle{plain}
\fi
\maketitle

\begin{abstract}

We propose a ConvNet model for predicting 2D human body poses in an image. The model regresses a heatmap representation for each body keypoint, and is able to learn and represent both the part appearances and the context of the part configuration.

We make the following three contributions: (i) an architecture combining a feed forward module with a recurrent module, where the recurrent module can be run iteratively to improve the performance; (ii) the model can be trained end-to-end and from scratch, with auxiliary losses incorporated to improve performance; (iii) we investigate whether keypoint visibility can also be predicted.

The model is evaluated on two benchmark datasets. The result is a simple architecture that achieves performance on par with the state of the art, but without the complexity of a graphical model stage (or layers). 

\end{abstract}

\section{Introduction}

Estimating 2D human poses from images is a challenging task with many applications in computer vision, such as motion capture, sign language and activity recognition. For many years approaches have used variations on the pictorial structure model~\cite{Felzenszwalb05,Fischler73} of a combination of part detectors and configuration constraints~ \cite{Andriluka09,Ferrari08,Pishchulin12,Sapp10,Yang11}. However, the advent of Convolutional Neural Networks (ConvNets), together with large scale training sets, has led to models that perform well in demanding scenarios with unconstrained body postures and large appearance variations~\cite{Chen14,Insafutdinov16,Tompson15,Toshev14}. As the individual part detectors, e.g.\ the hand and limb detectors, and the pairwise part detectors have become stronger, so the importance of the configuration constraints has begun to wane, with quite recent methods not even including an explicit graphical model~\cite{Belagiannis15,Carreira15,Pfister15a}.

\begin{figure*}[t]
\centering
\begin{tabular}{ccccccc}
{\tiny Input image}&{\tiny Auxiliary Loss}&{\tiny Iteration 0}&{\tiny Iteration 1}&{\tiny Iteration 2}&
{\tiny Iteration 3}&{\tiny Output}\\
\includegraphics[scale=0.215, angle=-0]{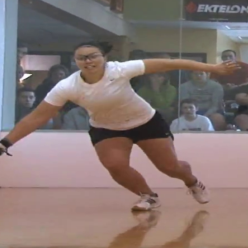}&
\includegraphics[scale=0.215, angle=-0]{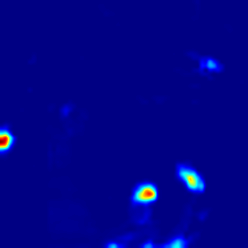}&
\includegraphics[scale=0.215, angle=-0]{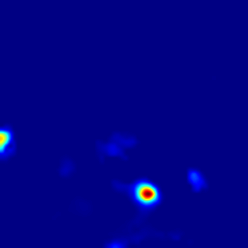}&
\includegraphics[scale=0.215, angle=-0]{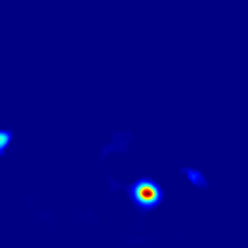}&
\includegraphics[scale=0.215, angle=-0]{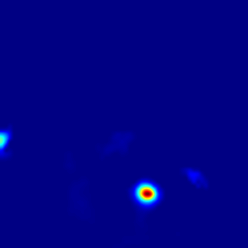}&
\includegraphics[scale=0.215, angle=-0]{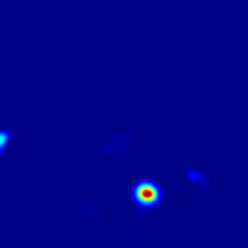}&
\includegraphics[scale=0.215, angle=-0]{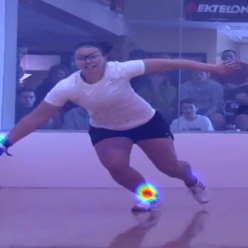}\\
\includegraphics[scale=0.215, angle=-0]{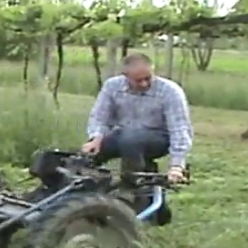}&
\includegraphics[scale=0.215, angle=-0]{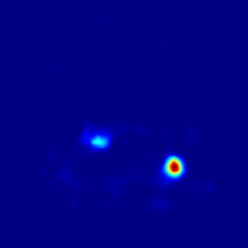}&
\includegraphics[scale=0.215, angle=-0]{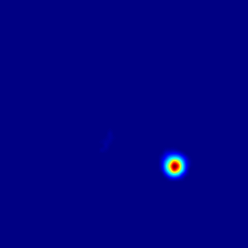}&
\includegraphics[scale=0.215, angle=-0]{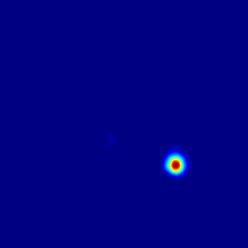}&
\includegraphics[scale=0.215, angle=-0]{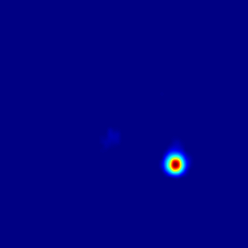}&
\includegraphics[scale=0.215, angle=-0]{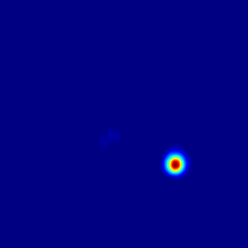}&
\includegraphics[scale=0.215, angle=-0]{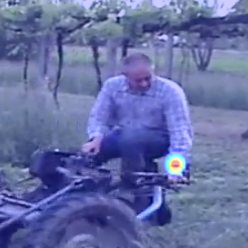}\\
\includegraphics[scale=0.215, angle=-0]{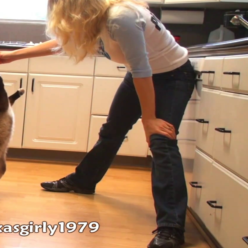}&
\includegraphics[scale=0.215, angle=-0]{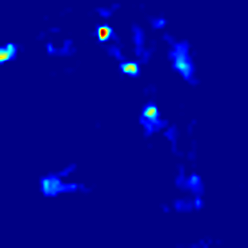}&
\includegraphics[scale=0.215, angle=-0]{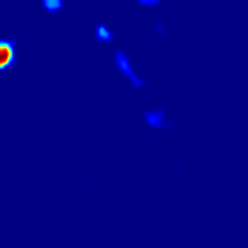}&
\includegraphics[scale=0.215, angle=-0]{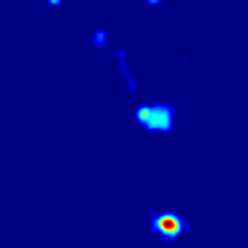}&
\includegraphics[scale=0.215, angle=-0]{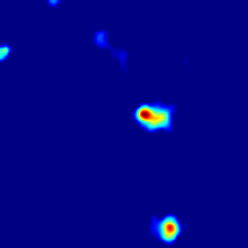}&
\includegraphics[scale=0.215, angle=-0]{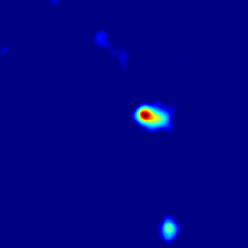}&
\includegraphics[scale=0.215, angle=-0]{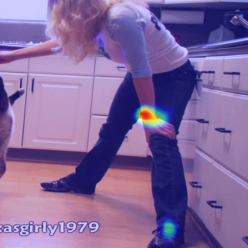}\\
\end{tabular}
\caption{\small {\bf Results of the Recurrent Human Model}: The predicted heatmaps (MPII Human Pose dataset~\cite{Andriluka14}) are visualized after every iteration of the recurrent module for the right ankle (\textit{first row}), left wrist (\textit{second row}) and left wrist (\textit{third row}). The model progressively suppresses false positive detections that occur at the first iterations.}
\label{fig:teaser}
\end{figure*}

\begin{figure*}[t]
\centering
\begin{tabular}{ccccccc}
{\tiny Input image}&{\tiny Auxiliary Loss - 1}&{\tiny Iteration 0}&{\tiny Iteration 1}&{\tiny Iteration 2}&
{\tiny Iteration 3}&{\tiny Output}\\
\includegraphics[scale=0.195, angle=-0]{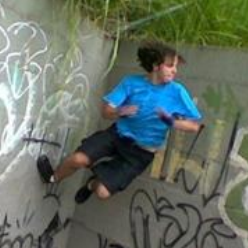}&
\includegraphics[scale=0.195, angle=-0]{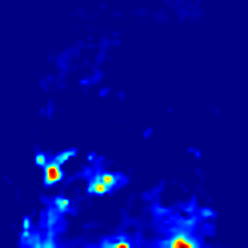}&
\includegraphics[scale=0.195, angle=-0]{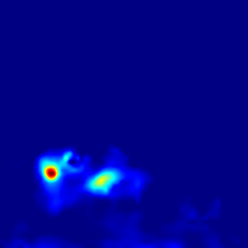}&
\includegraphics[scale=0.195, angle=-0]{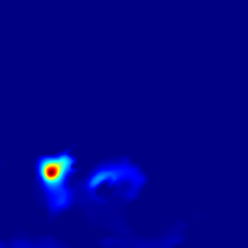}&
\includegraphics[scale=0.195, angle=-0]{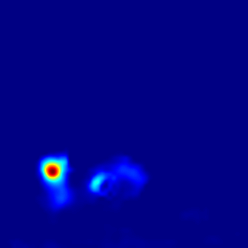}&
\includegraphics[scale=0.195, angle=-0]{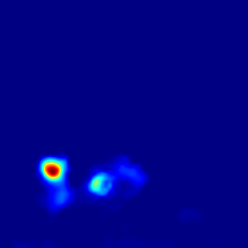}&
\includegraphics[scale=0.195, angle=-0]{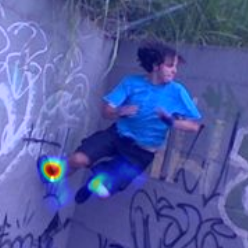}\\
\includegraphics[scale=0.195, angle=-0]{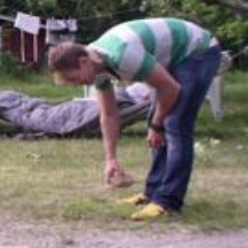}&
\includegraphics[scale=0.195, angle=-0]{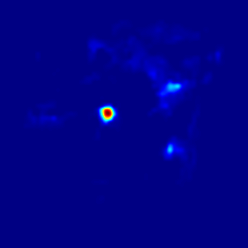}&
\includegraphics[scale=0.195, angle=-0]{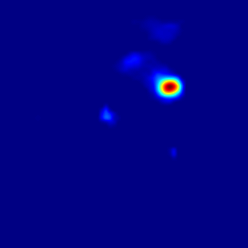}&
\includegraphics[scale=0.195, angle=-0]{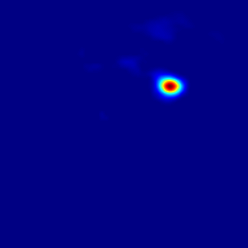}&
\includegraphics[scale=0.195, angle=-0]{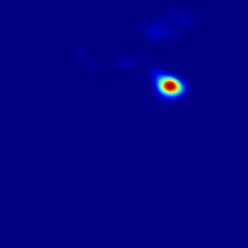}&
\includegraphics[scale=0.195, angle=-0]{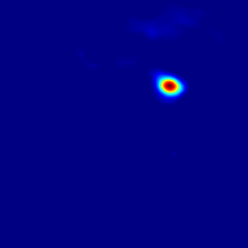}&
\includegraphics[scale=0.195, angle=-0]{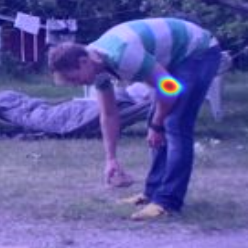}\\
\includegraphics[scale=0.195, angle=-0]{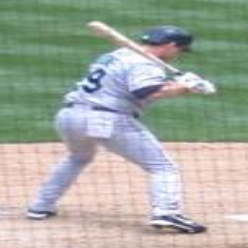}&
\includegraphics[scale=0.195, angle=-0]{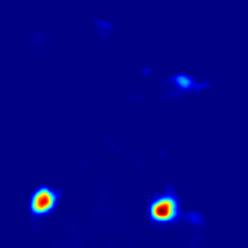}&
\includegraphics[scale=0.195, angle=-0]{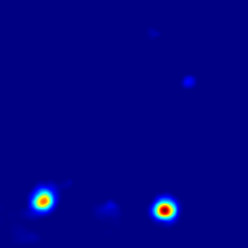}&
\includegraphics[scale=0.195, angle=-0]{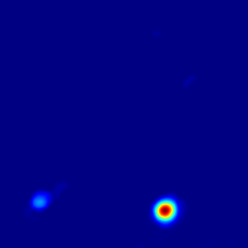}&
\includegraphics[scale=0.195, angle=-0]{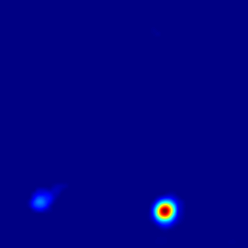}&
\includegraphics[scale=0.195, angle=-0]{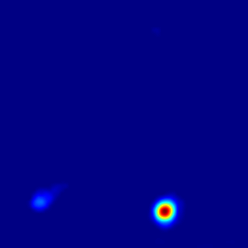}&
\includegraphics[scale=0.195, angle=-0]{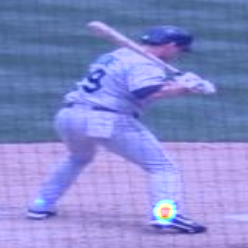}\\
\end{tabular}
\caption{\small {\bf More Results from the LSP dataset}: We visualize the predicted heatmaps after every iteration of the recurrent module for the right ankle (\textit{first row}), left elbow (\textit{second row}) and right ankle (\textit{third row}).}
\label{fig:teaser2}
\end{figure*}

In this paper, we describe a new ConvNet model and training scheme for human pose estimation that makes the following contributions: (i) a model combining a feed-forward module with a recurrent module, where the recurrent module can be run iteratively to increase the effective receptive field of the network and thus improve the performance (see Fig.~\ref{fig:teaser} and~\ref{fig:teaser2}); (ii) the model can be trained end-to-end, and auxiliary losses can be incorporated to improve performance; and (iii) a preliminary investigation into improving occlusion prediction in human pose estimation.

Our model is mainly inspired by two recent papers: Pfister {\it et al.}~\cite{Pfister15a} and Carreira {\it et al.}~\cite{Carreira15}. The first introduced the idea of `fusion layers', convolutional layers that {\em implicitly} encode a configuration model and capture context. The second introduced an iterative update module which progressively makes incremental improvement to the pose estimate. We borrow the fusion layers idea from~\cite{Pfister15a}, but apply it multiple times as a recurrent network in the manner of~\cite{Carreira15}. However, unlike~\cite{Carreira15} our model is trained end-to-end and does not require a rendering function for combining the output with the input.

\setlength{\tabcolsep}{0.3pt}
\begin{figure}[t]
\centering
\subfigure[{\small Keypoints}]{\label{fig:key}\includegraphics[scale=0.44, angle=-0]{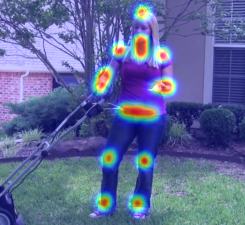}}
\subfigure[{\small Body parts}]{\label{fig:pair}\includegraphics[scale=0.44, angle=-0]{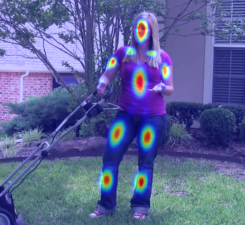}}\\
\subfigure[{\small Superimposed}]{\label{fig:super}\includegraphics[scale=0.44, angle=-0]{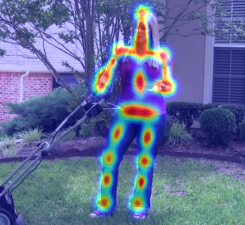}}
\subfigure[{\small Keypoint Prediction}]{\label{fig:pred}\includegraphics[scale=0.44, angle=-0]{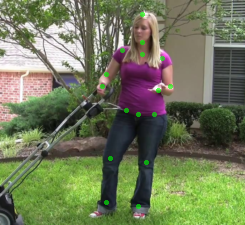}}\\
\caption{\small {\bf Regressed Heatmaps}: The regressed keypoint (a) and body part (b) heatmaps are presented for a validation sample. In (c), both heatmaps are superimposed, resulting in a human skeletal shape. The final outcome is the keypoint prediction (d), while the body part heatmaps act as an auxiliary task.}
\label{fig:heatmapsEx}
\end{figure}

In addition, our model shares
with Convolutional Pose Machines~\cite{Wei16} and the Hourglass
model~\cite{Newell16}
the motivation of using 
large convolution
kernels to capture more context (originally proposed by Pfister {\it
et al.}~\cite{Pfister15a}). Unlike these approaches, we use a
recurrent convolutional neural network to increase the receptive
fields which results in orders of magnitude less parameters in training.
Including the recurrent module multiple times is similar to the stacking 
of more hourglass modules in~\cite{Newell16}.

\setlength{\tabcolsep}{0.3pt}

The outcome of our approach is a simple recurrent model that reaches state-of-the-art performance on different standard benchmark datasets, but does not employ an explicit configuration model~\cite{Chen14} nor a complicated network architecture~\cite{Tompson15}.

\subsection{Related Work}\label{relatedWork}

For many years the `workhorse' in human pose estimation has been a tree structured graphical model, often based on the efficient pictorial structure methods of Felzenszwalb and Huttenlocher~\cite{Felzenszwalb05}. This supported a host of methods, including~\cite{Buehler11,Eichner12,Sapp10,Yang11}. An alternative approach,
which also included configuration constraints, was based around the poselet idea~\cite{Bourdev09,Gkioxari14}.

Early methods using ConvNets predicted pose coordinates of human keypoints directly (as 2D coordinates)~\cite{Toshev14}. An alternative, which it turns out might be better suited to ConvNets, is an indirect prediction by first regressing a heatmap over the image for each keypoint, and then obtaining the keypoint position as a mode in this heatmap~\cite{Jain14,Pfister15a,Tompson14,Tompson15}. The advantage of the heatmap over direct prediction is threefold: it mostly avoids problems with ConvNets predicting real values; it can handle multiple instances in the image (e.g.\ if there are several hands present and consequently several corresponding hand keypoints); and it can represent uncertainty by multiple modes.

Furthermore, combining heatmaps with large convolutional kernels and
deeper
models~\cite{Bulat16,Lifshitz16,Newell16,Pfister15a,Wei16} improves performance 
-- since the effective receptive fields, and consequently the context
captured, is increased.
For example, Pfister {\it et al.}~\cite{Pfister15a} added several large
convolutions (e.g.\ $13 \times 13$ kernels). However, a disadvantage
is that this increases the number of parameters
and makes the optimization more difficult.
In our model, we employ a recurrent module
that essentially increases the effective receptive fields without
introducing additional parameters.

The method of Carreira {\it et al.}~\cite{Carreira15} is an interesting hybrid that switches between regressing direct pose coordinates (as the output of the iterated module) and using a heatmap as the input (to the iterated module). In this
respect it is similar to the architecture of~\cite{Oberweger15a} which also switches between direct pose coordinates and an image representation in an iterated module. In other related work, the iterated implicit configuration module of our model bears similarities to auto-context~\cite{Tu10} and the message-passing inference machines of~\cite{Ross11}.
\section{Recurrent Human Pose Model Architecture}\label{methodSec}

Our aim is to predict 2D human body pose from a single image, represented as a set of keypoints. In this section, we describe our ConvNet model that takes the image as input, and learns to regress a heatmap for each keypoint, where the location of the keypoint is obtained as a mode of the heatmap.

\begin{figure*}[]
\centering
\begin{tabular}{c}
\includegraphics[scale=0.31, angle=0]{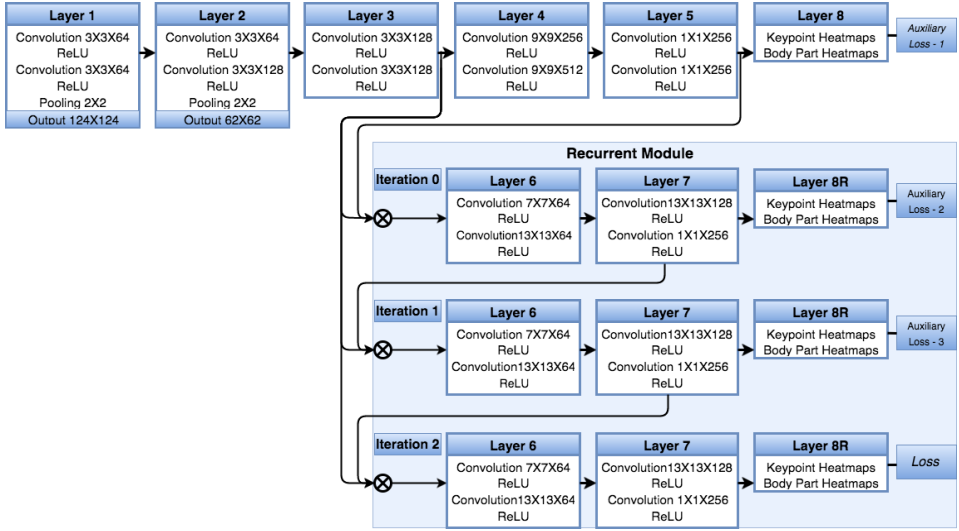} \label{fig:convergence}
\end{tabular}
\caption{\small {\bf ConvNet with Recurrent Module}: Our network is composed of $7$ layers. The recurrent model is introduced for Layer $6$ and $7$. In the current example, a network with $2$ iterations is visualized. Note that all loss functions are auxiliary for facilitating the optimization and the final outcome  comes from Layer 8D. Moreover, the body part heatmaps is an auxiliary task for additional data augmentation. In our graph, the symbol $\bigotimes$ corresponds to the concatenation operation.}
\label{fig:RCNNmodel}
\end{figure*}

The architecture of the ConvNet is overviewed in Fig.~\ref{fig:RCNNmodel}. It consists of two modules: a {\em Feed-Forward module} that is run once, and a {\em Recurrent module} that can be run multiple times. Both modules output heatmaps, and can be trained with auxiliary losses. However, the key design idea of the architecture is how context is apportioned in training and inference. The Feed-Forward module mainly acts as an independent `part' detector, regressing the keypoint heatmaps, but largely unaware of context from the configuration of other parts, due to the smaller effective receptive fields. In contrast, the recurrent module progressively brings in more context each time it is run, in part because the effective receptive field is increased with each iteration (Fig.~\ref{fig:teaser}).

In the following we describe the architecture of the two modules and the loss function used for training. The entire network can be trained end-to-end, but we also describe the use of auxiliary losses that can be employed to speed up the training and improve performance. We also investigate two other aspects: the benefit of including additional supervision in the form
of hallucinated annotation; and the benefit of training that is occlusion-aware.

\setlength{\tabcolsep}{0.3pt}

\subsection{Feed-forward Module}

The module is based on the heatmap regression architecture of~\cite{Pfister15a} with modifications. We use smaller filters (i.e.\ $3\times 3$) for the initial convolutional layers, combined with non-linear activations (Layer $1-3$ in Fig.~\ref{fig:RCNNmodel}). This idea from~\cite{Simonyan15} allows more non-linearities to be included in the architecture, and leads to better performance. Pooling is applied only twice in order to retain the output heatmap resolution sufficient large. The activation function is ReLU after every convolution and the prediction layers (Layer 8) are also convolutions, followed by ReLU, that output the predicted heatmaps.

From Layer $4$ to $6$, larger convolutions filters are employed to learn more of the body structure, followed by convolutions with $1\times 1$ filters (Layer $5$ and $7$). The skip layer concatenates the output from Layer $3$ and $5$, which composes the input for the fusion layer~\cite{Pfister15a} (Layer $6$ and $7$).

\subsection{Recurrent Module}

Our objective is to combine intermediate feature representations for learning context information and improving the final heatmap predictions. To that end, we introduce the recurrent module for the Layer $6$ and $7$ of our network. The input to the recurrent module is the concatenated output of Layer $3$ and Layer $7$. At every iteration, the input from Layer $3$ is fixed, while Layer $7$ is updated (see Fig.~\ref{fig:RCNNmodel}). Note that by using intermediate network layers for the recurrent module, we do not blend the predictions with the input, as in~\cite{Carreira15}. Finally, our network can be trained in an
end-to-end fashion.

\subsection{Body Part Heatmaps as Supplementary Supervision}

Inspired by the idea of part-based models~\cite{Andriluka09,Belagiannis14}, we
additionally propose body part heatmaps which are constructed by pairs of keypoints. In practice, we define the body part heatmap by taking the midpoint between the two keypoint as the center of the Gaussian distribution and define the variance based on the Euclidean distance between the two keypoints. Eventually, we model heatmaps for the body limbs, as it is depicted by Fig.~\ref{fig:heatmapsEx}. The keypoint heatmaps mostly represent body joints and the body part heatmap mainly capture limbs. Although, our main objective is to predict keypoints, modelling pairs of keypoints helps to capture additional body constraints and mainly acts as data augmentation, in terms of labels.

\setlength{\tabcolsep}{0.3pt}
\begin{figure*}[t]
\centering
\begin{tabular}{ccccccc}
{\tiny Input image}&{\tiny Auxiliary Loss - 1}&{\tiny Iteration 0}&{\tiny Iteration 1}&{\tiny Iteration 2}&
{\tiny Iteration 3}&{\tiny Output}\\
\includegraphics[scale=0.25, angle=-0]{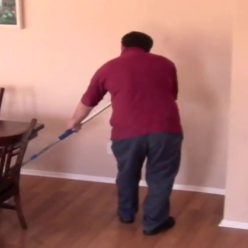}&
\includegraphics[scale=0.25, angle=-0]{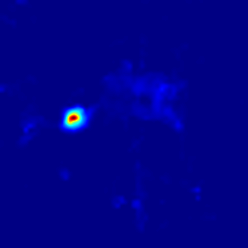}&
\includegraphics[scale=0.25, angle=-0]{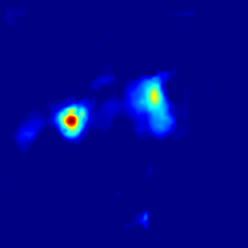}&
\includegraphics[scale=0.25, angle=-0]{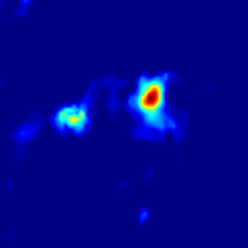}&
\includegraphics[scale=0.25, angle=-0]{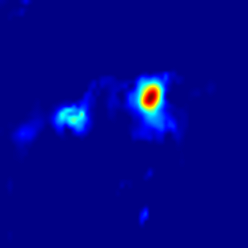}&
\includegraphics[scale=0.25, angle=-0]{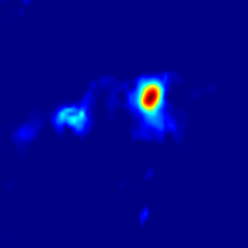}&
\includegraphics[scale=0.25, angle=-0]{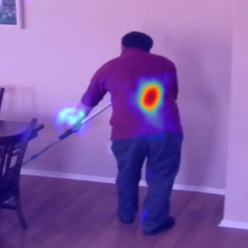}\\
\end{tabular}
\caption{\small {\bf Prediction of Occluded Keypoints}: 
The right wrist is erroneously predicted even though it is not visible
because the model learns to capture context and it accordingly
predicts. Note that occluded keypoints are not provided during
training as ground-truth labels, instead they are ignored or penalized
based on the occlusion scenario.}
\label{fig:occlFig}
\end{figure*}

\subsection{Target Heatmaps and Loss Function}

At training time, the ground-truth labels are heatmaps synthesised for each keypoint separately by placing a Gaussian with fixed variance at the ground truth keypoint position. We then use the mean squared error, which penalises the squared pixel-wise differences between the predicted heatmap and the synthesised ground-truth heatmap. 

The same loss is also used for the feed-forward part and the recurent module of the network. At every loss layer, we equally weights the keypoint and body part heatmaps. Finally, the training of the ConvNet is accomplished using backpropagation~\cite{Rumelhart86} and stochastic gradient descent~\cite{Bottou10}.

During training, Layer 8A is used as an auxiliary loss to comfort the optimization~\cite{Szegedy15}. In addition, we propose to use an auxiliary loss function at the end of every iteration of the recurrent module, other than the last iteration, to boost the gradients' magnitude during backpropagation. As a result, Layer 8B and 8C are auxiliary tasks and the actual prediction is the outcome of Layer 8D, given a network of 2 iterations as in Fig.~\ref{fig:RCNNmodel}. Finally, the cost function of our model for a set of $S$ training samples is defined as:
\begin{equation}
E=\sum_{s=1}^{S}\| \mathbf{h}^{s} - f( \mathbf{x},t; \boldsymbol{\theta})^{s}\|^{2}_{},
\end{equation}
where $\mathbf{h}^{s} $ is the synthesised ground-truth heatmap and $f(.)$ represents the ConvNet with learned parameters $\boldsymbol{\theta}$ and the recurrent module at the $t$ iteration.

\subsection{Occluded Keypoints} \label{SecOcclKey}

One of the most challenging aspects of predicting human body parts in images is dealing with the problem of occlusion -- both self-occlusion and occlusion by other entities.

With the context carried by the recurrent module, we have a new way of approaching this problem. A body keypoint, such as a wrist, generates a strong response in a heatmap for two reasons: because the keypoint is visible and because it can be inferred from the configuration of the other keypoints of the body. The latter can potentially be a problem if configuration dominates over visibility, and a keypoint is predicted even though it is occluded (Fig.~\ref{fig:occlFig}).

To this end, we investigate three different training scenarios for the
network: one {\em ignoring} occluded keypoints and body parts in the
loss function (since they are not visible), two {\em including}
occluded keypoints and body parts in the loss function to increase the
amount of training data, and three {\em excluding} occluded keypoints
and body parts by considering them as background and consequently
penalizing a corresponding heatmap response. In the first
scenario, heatmaps are synthesized, but the gradient values of
occluded keypoints or body parts are ignored at the respective heatmap
regions during backpropagation (i.e.\ zero gradient). Thus, the network
is trained without including occluded keypoints and at
the same time without penalizing them. In the second scenario, 
the occluded keypoints are included and we learn to infer them. In the third
scenario, the gradient values of occluded keypoints or body parts
encourage the heatmap areas of the occluded parts to converge to
zero. This penalizes heatmaps that erroneously infer occluded
parts at points predicted by context. Fortunately, the MPII Human Pose
dataset~\cite{Andriluka14} and LSP~\cite{Johnson11} datasets, which
are used for training, provide occluded keypoint annotation within the
context of predicted positions that can be used for this purpose.

\section{Implementation Details}

The network takes as input an RGB image with resolution $248\times 248$ and outputs heatmaps with resolution a quarter of the input that is $62\times 62$. The input image is normalized by mean subtraction at each channel. Furthermore, data augmentation is performed by rotating, scaling, flipping and cropping the input image. Regarding the network
parameters, the learning rate is set to $10^{-5}$ and gradually decreased to $10^{-6}$, while we found that no more than $40$ training epochs are required for obtaining a stable solution. Note that we train the model from scratch and the training time is less than $2$ days. The momentum is set to $0.95$ and the batch size to $20$ samples. Also, batch normalization~\cite{Ioffe15} is used for every convolutional layer other than the layers with $1\times 1$ filters and the output layers.

The generated target heatmaps have $\sigma$ variance set to $1.3$ for the keypoint Gaussian distributions, while the body part heatmaps have different variance for the $x$ and $y$ direction based on the Euclidean distance between the two keypoints that form a part. In particular, we set $\sigma_{x}$ and $\sigma_{y}$ equal to $0.15$ and $0.1$ of the Euclidean distance. Moreover, we found it crucial to weight the gradients of the heatmaps, since the heatmap data is unbalanced. A heatmap has most of its area equal to zero (background) and only a small portion of it corresponds to the Gaussian distribution (foreground). For that reason, it is important to weight the gradient responses so that there is an equal contribution to the parameter update between the foreground and background heatmap pixels. Otherwise, there is a prior towards the background that forces the network to converge to zero. In addition, we magnify the Gaussian distributions so that their mode is around to $12$ and consequently enlarge the difference between foreground and background pixels.

Furthermore a heatmap can include multiple individuals (e.g. MPII Human Pose dataset~\cite{Andriluka14}). For our experiments, it is assumed that one is the active individual and the predictions of the rest are ignored during backpropagation. As a result the network learns to predict a single body configuration.

The implementation of our model is in MatConvNet~\cite{Vedaldi15} and our code is publicly available\footnote{\url{http://www.robots.ox.ac.uk/~vgg/software/keypoint_detection/}}. In the next section, we evaluate the components of the  recurrent human model, examine how well the regressed heatmaps address the problem of occlusion detection and compare our results with related approaches.

\setlength{\tabcolsep}{0.3pt}

\section{Experiments}
We evaluate the components of our model and compare with related methods for the task of 2D human pose estimation from a single image. The evaluation is based on the MPII Human Pose~\cite{Andriluka14} and LSP~\cite{Johnson11} datasets. On the MPII Human Pose~\cite{Andriluka14} dataset, we evaluate for single human pose estimation, while the LSP~\cite{Johnson11} dataset includes labels only for single human evaluation. Keypoint annotation is provided for both datasets, $16$ keypoints in MPII Human Pose (Fig.~\ref{fig:heatmapsEx}) and $14$ in LSP, which we use for generating the target ground-truth keypoint and body part heatmaps for training. The parameters of the model, such as the learning rate and number of training epochs, are defined based on the validation dataset of MPII Human Pose, as proposed by~\cite{Tompson15}, and remain the same for
all evaluations. Moreover, the validation dataset of~\cite{Tompson15} is used for all the baseline evaluations. Our network architecture is significantly different from the common recognition models~\cite{Krizhevsky12,Simonyan15} and thus we choose to train from scratch instead of fine-tuning  a pre-learnt model.

\begin{figure}[]
\centering
\begin{tabular}{c}
\includegraphics[scale=0.42, angle=-0]{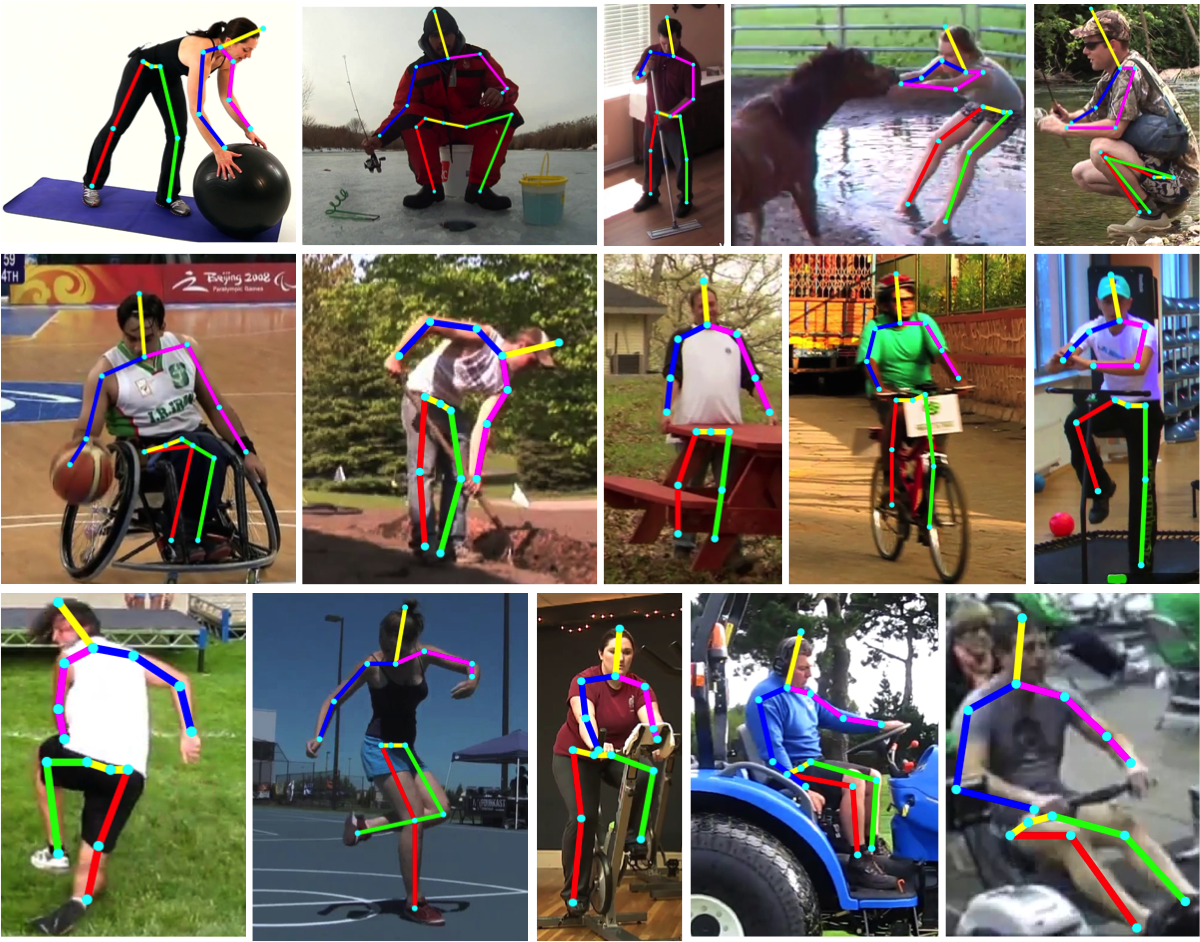}
\end{tabular}
\caption{\small {\bf Pose Results on MPII Human Pose}: the predictions are
from two iterations of the recurrent module.}
\label{fig:resultsPoseMPI}
\end{figure}

The evaluation of the recurrent human model is divided into three parts: the component evaluation, occlusion evaluation and comparison with related methods. The different parts of our model are examined in the model components evaluation. In the occlusion part, we evaluate the potential of our model to predict whether a keypoint is visible. Finally, we compare our model with related methods, mainly deep learning approaches. The main performance metric for the evaluations is the PCKh measure~\cite{Andriluka14}. Based on the PCKh definition, a keypoint is correctly localized if the distance between the predicted and ground-truth keypoint is smaller than $50\%$ of the head length. 

\subsection{Component Evaluation}

The proposed model is composed of different objectives and the recurrent module, where the recurrent module can include several iterations. In this evaluation, we investigate the contribution of each component to the final performance. For this purpose, we rely on the MPII Human Pose~\cite{Andriluka14} dataset with the validation dataset from~\cite{Tompson15}. The results of the component evaluation are summarized in Table~\ref{compoEval}.

\begin{table}[h]
\centering
\caption{{\small {\bf Model Components}}: We evaluate the components of the model for the body keypoints on MPII Human Pose dataset using the PCKh metric and the validation dataset from~\cite{Tompson15}. First, we evaluate our model by training under different occlusion settings (including, ignoring or excluding occluded keypoints). Next, different number of recurrent iterations is examined and finally the body part heatmaps are added to the model. We also report results by training an equivalent model (2 iter. and body parts) in MSCOCO dataset~\cite{Lin214} and then fine-tuning on the MPII dataset. At the end, we investigate the model's performance by adding scale augmentation during testing.}
\label{compoEval}
\begin{tabular}{l|c|c|c|c|c|c|c|c}
Heatmaps                 & Head & Shoulder & Elbow & Wirst & Hip & Knee & Ankle & PCKh \\ \hline
Keypoint (exclude occl.)      &95.4  & 89.6  & 79.1  & 74.3  & 78.9  & 73.0 & 66.7 &  80.3 			    \\ \hline
Keypoint (ignore occl.)        &95.3  & 91.4  & 81.9  & 75.2  & 80.5  & 73.1 & 67.4 &  81.4 			    \\ \hline
Keypoint (include occl.)       &95.2  & 92.2  & 82.9  & 77.0  & 82.5  & 75.7 & 69.6 &  82.8 			    \\ \hline 
\quad + 1 Iteration         &96.1  & 93.0  & 84.0  & 77.5  & 83.5  & 76.2 & 69.7 & 83.6						   \\ \hline
\quad + 2 Iterations        &96.1  & 93.0  & 84.1  & 77.4  & 83.4  & 76.1 & 70.0 & 83.6						   \\ \hline
\quad \quad + Body Part &96.1  & 93.1  & 84.9  & 78.3  & 84.6  & 78.5 & 72.6 & 84.6    \\ \hline
\quad \quad \quad + Fine-tune& 96.3  & 94.0  & 85.9  & 80.2  & 86.0  & 80.0 & 75.7& 86.0		     \\\hline
\quad \quad \quad \quad + Scale Aug.& 96.4  & 94.0  & 86.3  & 80.2  & 86.4  & 80.5 & 75.9& 86.3		     \\\hline \hline
\hline
\end{tabular}
\end{table}

At first, we evaluate the objective of the keypoint heatmaps, w/o occluded keypoints (first three rows, Table~\ref{compoEval}). Including occluded points during training (i.e. constructing heatmaps using them) gives the best performance because of the larger amount of training data. This evaluation composes the baseline of the proposed model. Next, recurrent iterations are added to the keypoint model. Table~\ref{compoEval} shows that one iteration is sufficient and adding more does not improve the final result. To gain more from the recurrent model, we add the objective of the body parts (sixth row, Table~\ref{compoEval}). We do not aim to predict body parts, but observe that this additional objective is helpful for capturing additional body constraints, propagate back more gradients and thus it brings a boost to the model's performance. We notice that after two recurrent iterations, there is not significant improvement of the final result. In general, the parts that are already well predicted using the feed-forward model benefit less from the recurrent module (e.g. head). Note that the model with one iteration that also includes body parts has the same performance as the model with two iterations and only keypoints objective. To examine how additional amount of training data affects model's behaviour, we included the MSCOCO dataset~\cite{Lin214} during training, but there was not improvement. On the hand, training first our full model (i.e. 2 iterations and body parts) on MSCOCO dataset and then fine-tuning it on MPII dataset improved our final performance (seventh row, Table~\ref{compoEval}). Finally, scale augmentation is added at test time to gain another small boost in our performance.

\subsection{Occlusion Prediction}

In this experiment, we analyse the potential of the heatmaps to
predict the visibility of a keypoint. Empirically, the heatmaps of
occluded keypoints tend to have low responses (in terms of magnitude);
and, as a result, the visibility of a keypoint can be inferred from
the heatmap responses. We evaluate to what extent the response
magnitude can be used to predict keypoint visibility on the validation
set of the MPII Human Pose dataset, since this dataset provides
occlusion labels. However, the distribution of visible and occluded
keypoints is unbalanced (only $\sim 23\%$ of annotated keypoints are
tagged as occluded), so there is bias towards the visible keypoints.

For the evaluation we make the assumption that there should only be a
single response for each heat map for a visible point (and no or a low
response if the point is occluded) since the data set has only one of
each keypoint (e.g.\ left elbow) for each test sample. We then pick
the maximum response in each heatmap, and order all these maximum
responses (over all images and all heatmaps) by their strength.  A
Precision-Recall curve is then computed where the positives are
visible points (and negatives are the occluded points).  Note, since
the evaluation examines only the heatmap responses, and not their
positions, we are not determining whether the predicted visible point
is at the correct position or not.

\begin{figure}[]
\centering
\begin{tabular}{c}
\includegraphics[scale=0.12, angle=-0]{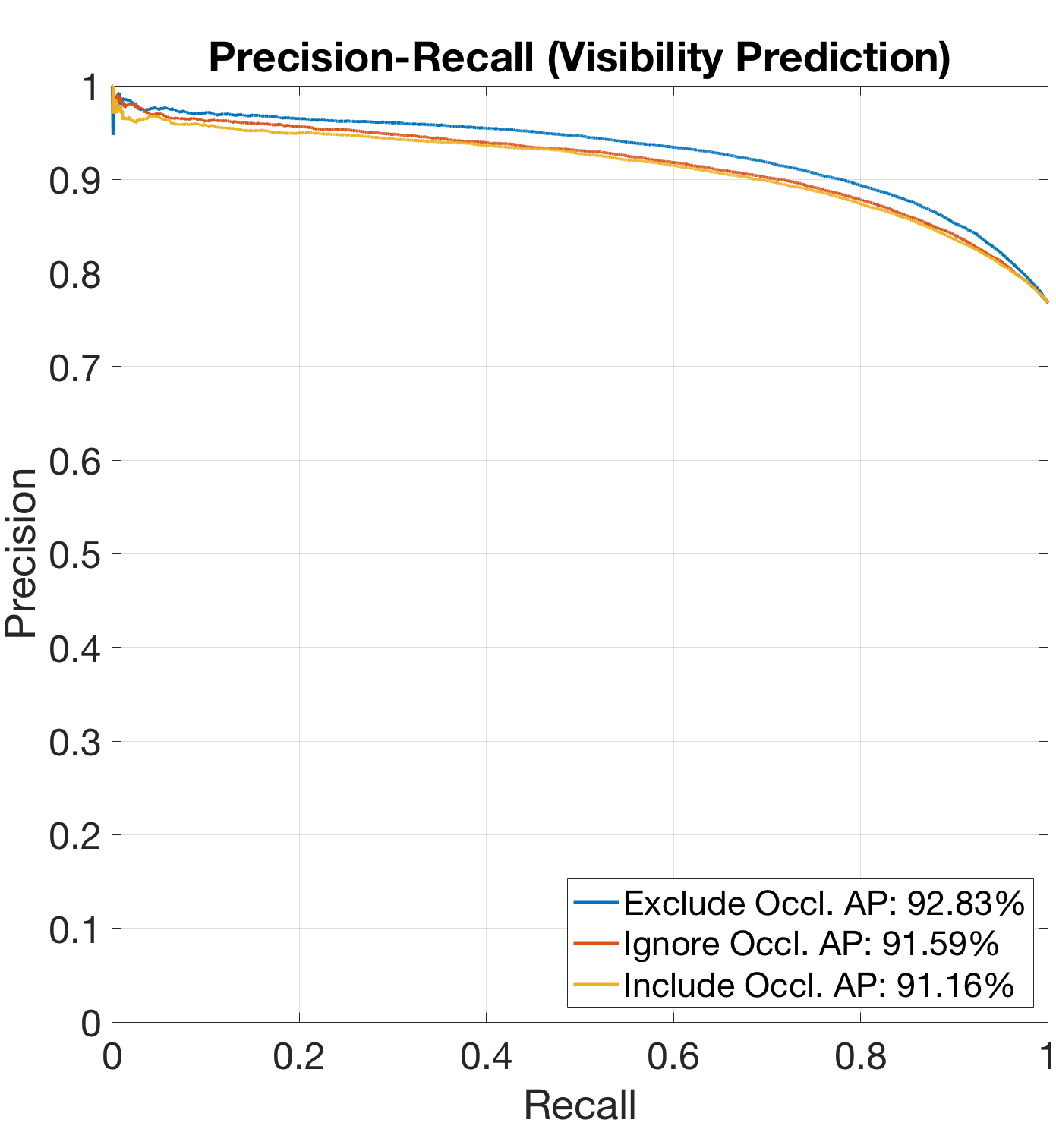}\\
\end{tabular}
\caption{\small {\bf Visibility Prediction Precision-Recall}: The evaluation is performed using three different models: the first model is trained by ignoring occluded keypoints and body parts (\textit{Ignore Occl.}) during training, the second by including them (\textit{Include Occl.}) to increase the amount positive training data and the last by excluding (\textit{Exclude Occl.}) them and thus treating them as background. The average precision (AP) is reported for all training scenarios.}
\label{fig:resultsOcc}
\end{figure}

\begin{figure}[]
\centering
\subfigure[{\footnotesize Include Occl. Right Ankle}]{\label{fig:occ1}\includegraphics[scale=0.33, angle=-0]{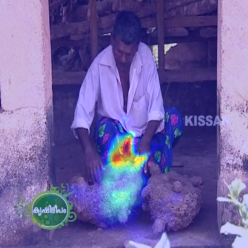}}
\subfigure[{\footnotesize Exclude Occl. Right Ankle}]{\label{fig:occ2}\includegraphics[scale=0.33, angle=-0]{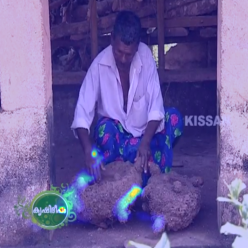}}\\
\subfigure[{\footnotesize Include Occl. Left Elbow}]{\label{fig:occ3}\includegraphics[scale=0.34, angle=-0]{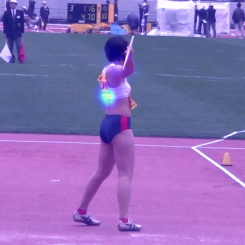}}
\subfigure[{\footnotesize Exclude Occl. Left Elbow}]{\label{fig:occ4}\includegraphics[scale=0.34, angle=-0]{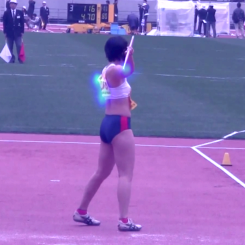}}
\caption{\small {\bf Visibility Heatmaps}: 
On~\ref{fig:occ1}, the predicted \textit{right ankle} heatmap is visualized for the model that includes the occluded keypoints during training, while on ~\ref{fig:occ2} the same heatmap is presented for the model that excludes (i.e. penalizes) the occluded keypoints (during training). Similarly on ~\ref{fig:occ3}, the self-occluded \textit{left elbow} is recovered due to the context information, while the response at the correct area is low in\ref{fig:occ4} for the model that has been trained by excluding during training occluded keypoints.}
\label{fig:occlHeatmap}
\end{figure}

This experiment is performed using the three training scenarios that
were defined in Sec.~\ref{SecOcclKey} using the model with 1
iteration, including body parts. In the first case, the occluded
keypoints and body parts are ignored from the ground-truth labels. In
the second case, the occluded keypoints and body parts are included to
the training, while in the third case, the heatmap regions of the
occluded keypoints are penalized (i.e.\ considered as background). Our
results are summarized in Fig.~\ref{fig:resultsOcc}. One can see that
the model with penalized occluded keypoints (in the training process)
performs better than the models that ignore or include occluded
keypoints and body parts. In practice, we observe that the network
learns to treat areas of occluded keypoints as background (see
Fig.~\ref{fig:occlHeatmap}). Nevertheless, we find that the overall
performance of the network that penalizes occluded keypoints during
training is around $3\%$ worse than the network that includes the
occluded keypoints (Table~\ref{compoEval}) -- this is a consequence of
the fact that the MPII evaluation ignores occluded keypoints, so there
is no disadvantage in predicting them, and they clearly provide some
context in training. Our average precision (AP) is more than $90\%$
for the case of the visibility prediction. 
We do not compare
with another approach since we are not aware of any related method
that performs occlusion detection on this dataset. 
Our results are not directly
comparable to the $35\%$ of average detection accuracy of occluded
joint from~\cite{Rafi15} or the $85\%$ of accuracy of occlusion
prediction from \cite{Chen2015}; but these evaluations are indicative
that our performance is good for this problem.

\subsection{Comparison with other Methods}

In our last experiment, we compare our results with related methods on the MPI Human Pose~\cite{Andriluka14} and LSP~\cite{Johnson11} datasets. In both evaluations, our model is executed for two iterations. We do not use any ground-truth information for the localization of the individuals in the LSP dataset, while rough localization is provided for the MPI Human Pose dataset.

\paragraph{MPII Human Pose Dataset} 
We use the same training and validation protocol
as~\cite{Tompson15}. Our results are summarized in
Table~\ref{tab:MPIcomp} and also samples are visualized in
Fig.~\ref{fig:resultsPoseMPI}. In all cases, we achieve on par
performance with other methods. It is worth noting that our model
architecture is significantly simpler than~\cite{Tompson15} and it
does not depend on a graphical model inference
as~\cite{Tompson14}. One should also observe that our model performs
better than the iterative method of Carreira {\it et
al.}~\cite{Carreira15} which relies on a pre-trained model and
training in stages.  In terms of the number of model parameters, as
can be seen from Table~\ref{table:MPIIparams}, our model has two
orders of magnitudes fewer parameters, and thus smaller capacity, than
the Hourglass model~\cite{Newell16}, and a third of the parameters 
of the Convolutional Pose
Machines~\cite{Wei16}, though it has comparable performance in the
evaluation.

\begin{table}[]
\centering
\caption{\small {\bf MPII Human Pose Evaluation}. The PCKh measure is used for the evaluation. The scores are reported for each keypoint separately and for the whole body. The area under the curve (AUC) is also reported. In addition, we include the results by training first our model on MSCOCO and then fine-tuning on the MPII dataset.}
\label{tab:MPIcomp}
\begin{tabular}{l|c|c|c|c|c|c|c|c|c}
 &Head & Shoulder & Elbow & Wrist & Hip & Knee & Ankle & Total & AUC\\
 Ours& 97.5  & 94.3  & 86.9  & 80.8  & 86.7  & 80.7 & 76.0 & 86.7 & 56.8\\
 Ours (fine-tuned)& 97.7  & 95.0  & 88.2  & 83.0  & 87.9  & 82.6 & 78.4 & 88.1 & 58.8\\\hline\hline
 Tompson et al.~\cite{Tompson14}& 95.8 & 90.3 & 80.5 & 74.3 & 77.6 & 69.7 & 62.8 & 79.6 & 51.8 \\
 Carreira et al.~\cite{Carreira15}& 95.7 & 91.7 & 81.7 & 72.4 & 82.8 & 73.2 & 66.4 & 81.3 & 49.1 \\
 Tompson et al.~\cite{Tompson15}& 96.1 & 91.9 & 83.9 & 77.8 & 80.9 & 72.3 & 64.8 & 82.0 & 54.9 \\
 Pishchulin et al.~\cite{Pishchulin16}& 94.1 & 90.2 & 83.4 & 77.3 & 82.6 & 75.7 & 68.6 & 82.4 & 56.5 \\ 
 {\scriptsize Insafutdinov} et al.~\cite{Insafutdinov16}& 96.8  & 95.2  & 89.3  & 84.4  & 88.4  & 83.4 & 78.0 & 88.5 & 60.8 \\
 Wei et al.~\cite{Wei16}& 97.8  & 95.0  & 88.7  & 84.0  & 88.4  & 82.8 & 79.4 & 88.5 & 61.4\\
 Bulat et al.~\cite{Bulat16}& 97.9  & 95.1  & 89.9  & 85.3  & 89.4  & 85.7 & 81.7 & 89.7 & 59.6 \\
 Newell et al.~\cite{Newell16}& 98.2  & 96.3  & 91.2  & 87.1  & 90.1  & 87.4 & 83.6 & 90.9 & 62.9 \\
\end{tabular}
\end{table}

\setlength{\tabcolsep}{0.3pt}

\begin{table}
\small
\caption{\small {\bf Model Size}. 
The number of convolutional parameters (i.e. number of input filter channels $\times$ filter height $\times$ filter width $\times$ output filter channels) for Wei {\it et al.}~\cite{Wei16} and Carreira {\it et al.}~\cite{Carreira15} is calculated from the online models; while the authors of~\cite{Newell16} have communicated the approximate number of convolutional and deconvolutional parameters in the Hourglass model. All examined model configurations are used for the MPII dataset evaluation.}
\begin{center}
    \begin{tabular}{l|r}
                  & Model Parameters \\ \hline
        Newell {\it et al.}~\cite{Newell16}     &$23.7 \times 10^{6}$ \\
        Wei {\it et al.}~\cite{Wei16}     & $29.7 \times 10^{6}$  \\
        Ours  & $15.4 \times 10^{6}$\\
        Carreira {\it et al.}~\cite{Carreira15}     & $10.0 \times 10^{6}$  \\
        \hline
    \end{tabular}
\end{center}
\label{table:MPIIparams}
\end{table}

\begin{figure}
\centering
\begin{tabular}{c}
\includegraphics[scale=0.45, angle=-0]{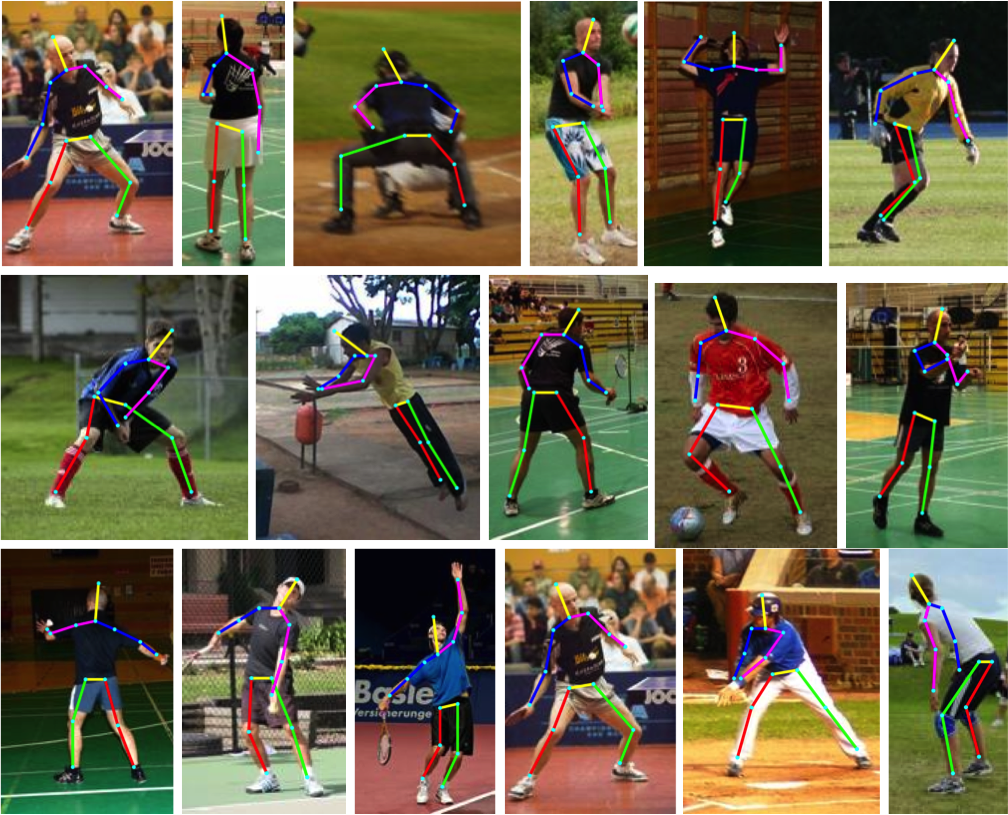}
\end{tabular}
\caption{\small {\bf Pose Results on LSP Dataset}: the result after two iterations in the recurrent module.} 
\label{fig:resultsPoseLSP}
\end{figure}

\paragraph{LSP Dataset} 
The dataset is composed of $2000$ images, where half of the images are used for training (Fig~\ref{fig:resultsPoseLSP}). There is also the extension of LSP~\cite{Johnson11} with $10000$ training samples which we use for this experiment. However, we observe that the training data is not sufficient for training our model from scratch, and thus we merge the training data of the extended LSP with the MPII Human Pose dataset. We also report results using a model trained or fined-tuned on the MPII Human Pose dataset. Our results are presented in Table~\ref{tab:LSPcomp}. The evaluation is accomplished using the PCK measure (threshold at 0.2) that is similar to PCKh, but it has 
as reference part the length of the torso instead of the head. It is clear
that we achieve promising performance for all keypoints. In
particular, our recurrent human model performs better than the
iterative method of Carreira {\it et al.}~\cite{Carreira15}, as well
as, the graph-based model with deep body part detectors of Chen \&
Yuille~\cite{Chen14}.

\begin{table}[]
\centering
\caption{\small {\bf LSP Evaluation}. The PCK measure is used for the evaluation. The scores are reported for each keypoint separately and for the whole body. We report results using the trained model from MPII Human Pose, the MPII model fined tuned on the extended LSP training data, and also training a new model by combining the training data of the MPII Human Pose with the extended LSP dataset.}
\label{tab:LSPcomp}
\begin{tabular}{l|c|c|c|c|c|c|c|c}
&Head & Shoulder & Elbow & Wrist & Hip & Knee & Ankle & Total\\

Ours (MPII model) & 90.8  & 84.4  & 76.3  & 70.4  & 81.5  & 81.9 & 77.8 & 80.5 \\
Ours (MPII fine-tuned)&94.3  & 87.1  & 78.6  & 72.0  & 78.1  & 83.2 & 77.1 & 81.5 \\
Ours (MPII \& LSP, 1 it.)&95.6   &88.8 &80.7  &75.5  &83.0 &86.2 &80.6&84.3\\
Ours (MPII \& LSP, 2 it.)&95.2  & 89.0  & 81.5  & 77.0  & 83.7  & 87.0 & 82.8 & 85.2
\\\hline\hline Pishchulin et al.~\cite{Pishchulin13}&87.2
&56.7 &46.7 &38.0 &61.0 &57.5 &52.7 &57.1\\ Wang Li et
al.~\cite{Wang13e}& 84.7 &57.1 &43.7 &36.7 &56.7 &52.4 &50.8
&54.6\\ Carreira et al.~\cite{Carreira15}&90.5 &81.8 &65.8
&59.8 &81.6 &70.6 &62.0 &73.1\\ Chen \&
Yuille~\cite{Chen14} &91.8&78.2& 71.8& 65.5& 73.3& 70.2&
63.4& 73.4\\ Fan et al.~\cite{Fan15} &92.4 & 75.2& 65.3&
64.0& 75.7& 68.3 & 70.4& 73.0\\
Pishchulin et al.~\cite{Pishchulin16} &97.0	&91.0	&83.8	&78.1	&91.0	&86.7	&82.0	&87.1\\
Wei et al.~\cite{Wei16} & 97.8& 92.5&87.0&83.9&91.5&90.8&89.9&90.5\\
Bulat et al.~\cite{Bulat16} & 97.2 & 92.1 & 88.1 & 85.2 & 92.2 & 91.4 & 88.7 & 90.7
\end{tabular}
\end{table}

\section{Conclusion}

We have introduced a recurrent human model for 2D human pose
estimation that is able to capture context iteratively, resulting in
improved localization performance.  We demonstrate performance
comparable to the state-of-the-art on two challenging human pose
estimation datasets, training the model from scratch. Finally, the
regressed heatmaps can be useful for predicting occlusion of
keypoints.

Future work will investigate whether the heat map obtained by 
combining the keypoints and body parts
(shown in Fig.~\ref{fig:heatmapsEx}(c)) can be used to avoid erroneous
keypoint predictions (such as left/right hand swopping).

\section{Acknowledgements}
We are grateful for the help provided by Tomas Pfister. Funding for this research was provided by the EPSRC Programme Grant Seebibyte EP/M013774/1.

\bibliographystyle{ieee}
\bibliography{shortstrings,vgg_local,vgg_other}

\end{document}